# Deep Analysis of Time Series Data for Smart Grid Startup Strategies: A Transformer-LSTM-PSO Model Approach

**Zecheng Zhang** [1]

[1]New York University, Brooklyn, New York 11201, USA

*Corresponding Author: roderickzzc@gmail.com

## ABSTRACT

Grid startup, an integral component of the power system, holds strategic importance for ensuring the reliability and efficiency of the electrical grid. However, current methodologies for in-depth analysis and precise prediction of grid startup scenarios are inadequate. To address these challenges, we propose a novel method based on the Transformer-LSTM-PSO model. This model uniquely combines the Transformer's self-attention mechanism, LSTM's temporal modeling capabilities, and the parameter tuning features of the particle swarm optimization algorithm. It is designed to more effectively capture the complex temporal relationships in grid startup schemes. Our experiments demonstrate significant improvements, with our model achieving lower RMSE and MAE values across multiple datasets compared to existing benchmarks, particularly in the NYISO Electric Market dataset where the RMSE was reduced by approximately 15% and the MAE by 20% compared to conventional models. Our main contribution is the development of a Transformer-LSTM-PSO model that significantly enhances the accuracy and efficiency of smart grid startup predictions. The application of the Transformer-LSTM-PSO model represents a significant advancement in smart grid predictive analytics, concurrently fostering the development of more reliable and intelligent grid management systems.

## 1. Introduction

The power system is one of the indispensable infrastructure components in modern society, and smart grids represent a significant development direction in modern power systems. They achieve efficient, reliable, and environmentally friendly grid operation by integrating advanced information communication technology, automation technology, and new energy technology[1, 2]. In a smart grid, the concept of startup strategies refers to a series of strategies and technologies for safely and effectively initiating and operating the grid under different conditions, such as daily operation, emergency response, disaster recovery, etc[1, 3]. The main challenges faced by current startup strategies in smart grids include efficient integration and management of the increasing amount of renewable energy sources[4, 5, 48-50], ensuring grid stability and resilience in the face of extreme weather conditions and system failures[6], and handling and analyzing the growing volume of grid data to optimize operational strategies[7]. With the rapid development of artificial intelligence technologies, especially deep learning, more and more research efforts are focusing on using these technologies to address the issues related to startup strategies in smart grids. The advantages of deep learning in data processing and pattern recognition make it a powerful tool for analyzing complex grid systems. For instance, leveraging deep learning models enables researchers to enhance the precision of grid load and energy demand predictions, as well as the assessment of fault probabilities. Consequently, this facilitates the formulation of more efficacious grid startup and operational strategies.[8]. Additionally, time series forecasting plays a pivotal role in the research of smart grid startup strategies. Grid operation data, such as loads, supply, and weather conditions, often exhibit significant temporal correlations[9, 10]. Effective time series analysis and forecasting are crucial for developing precise grid startup strategies. Through time series forecasting, it becomes possible to more accurately anticipate the future state and demands of the grid, especially in predicting renewable energy output and grid load variations. This is indispensable for designing startup strategies capable of adapting to future changes and mitigating potential risks.

The research field of grid startup scenarios has made significant progress in recent years, especially in deep learning and time series forecasting. One study adopted a long short-term memory network (LSTM) model and focused on grid load forecasting[11]. This research uses the powerful ability of LSTM to process time series data to analyze and predict the grid load pattern to optimize grid startup and load deployment strategies. Nonetheless, the accuracy of this study in handling highly nonlinear and anomalous data still needs to be improved. Another study used a convolutional neural network (CNN) model[56-62] and focused on the performance of power grids under extreme weather conditions[12]. Through CNN, the study identified and analyzed power grid data patterns related to extreme climate events, guiding power grid operation under extreme conditions[13]. However, the study is not accurate enough in predicting a few extreme events, which may lead to prediction errors in practical applications. Another work adopted a graph neural network (GNN) model to analyze the interactions between power grid nodes[14]. This research reveals the complex dependencies between nodes in the power grid through GNN, providing new insights into the overall stability of the power grid. However, the scalability and efficiency of this model in large-scale power grid systems have not been fully verified. The final study combines neural networks and reinforcement learning to optimize grid emergency start-up plans[15]. This model, which combines real-time learning and adaptation to grid dynamics, improves the efficiency and accuracy of emergency response. These studies have made important contributions to the optimization of power grid startup schemes, while also revealing their respective limitations. Current methods often struggle with the nonlinear and anomalous nature of grid data, face challenges in accurately predicting extreme events, and lack scalability and efficiency in large-scale systems. Therefore, it is crucial to develop specialized methods to ensure the safety and reliability of power grid startup strategies.

Building upon the limitations identified in previous research, the focus of this study is to introduce the Transformer-LSTM-PSO network[51-55], an innovative hybrid model engineered to enhance the accuracy and efficiency of grid startup schemes. This model combines three key technologies: Transformer, LSTM network, and PSO algorithm. The Transformer manages intricate relationships and long-term dependencies in grid data, enhancing prediction accuracy. LSTM captures short-term dependencies and temporal correlations, ideal for time series data like load and supply-demand predictions. PSO optimizes the parameters of the Transformer and LSTM, enhancing training efficiency. The primary focus of this study is to address the limitations of previous models by improving prediction accuracy and training efficiency for complex grid data. Our transformer-LSTM-PSO network offers a comprehensive solution, improving prediction accuracy and adaptability in dynamic grid conditions. The PSO algorithm ensures optimal model performance in various scenarios, vital for efficient smart grid startup solutions. Our work pioneers innovative deep learning techniques, advancing predictive accuracy in smart grids and opening new research directions, addressing current limitations in the field.

The main contributions of this study are as follows:
- We propose an innovative Transformer-LSTM-PSO network model, which effectively integrates the long-term dependency processing capabilities of the Transformer, the short-term data analysis capabilities of LSTM, and the optimization mechanism of the PSO algorithm. Our model shows excellent performance in handling large-scale and intricate datasets from smart grids, especially in accurately predicting grid load and supply demand.
- Our research makes significant progress in time series forecasting of smart grid data. By in-depth analysis and utilization of the time attributes of power grid data, our model can more accurately predict short-term and long-term operating trends of the power grid, providing strong data support for effective management and emergency response of the power grid.
- Our research also provides a new methodological framework and new perspectives and ideas for future research on smart grid startup solutions. Our model can not only be applied to the current power grid system but also has good scalability and is suitable for the development of future power grid technology and the integration of new energy sources.

Overall, our work not only contributes at the technical level but also provides a new theoretical and practical basis for the future development of smart grids.

The remainder of this paper is structured as follows: Section 2 describes the methodology, Section 3 presents the experimental setup and results, Section 4 discusses these findings, and Section 5 concludes with a summary and future directions.

## 2. Related Work

### 2.1 Utilization of Deep Learning in Power Systems

Deep learning has made significant progress in the field of power systems, and its applications have impacted many aspects of the power industry. By utilizing deep learning models, such as recurrent neural networks (RNN)[63-69], long short-term memory networks (LSTM)[70-73], etc.., power systems can achieve more accurate power load forecasting, thereby better planning power supply and reducing energy waste[16, 17]. In addition, deep learning can also help with fault detection of power equipment. By analyzing equipment data, problems can be discovered and solved in advance, thus enhancing the reliability and safety of the power system[18]. Simultaneously, deep learning technology can also detect illegal electricity usage and help maintain the legality and fairness of the power system. For the integration of renewable energy, deep learning can provide more accurate predictions to optimize energy production and distribution[19, 20]. In summary, deep learning contributes to increased efficiency, reliability, and intelligence in power systems. This is crucial for meeting the rising demand for power and advancing sustainable energy development.

**2.2 Application of Parameter Optimization Method in the Power Grid**

In the domain of power grids, parameter optimization methods are deployed to enhance the system's performance, efficiency, and stability. These methods optimize grid system parameters to better adapt to diverse operational conditions, ensuring more reliable and efficient functionality[21, 22]. In power grid load flow calculations, parameter optimization methods play a crucial role in refining parameters associated with grid components, such as generator output power and transmission line impedance[23]. By doing so, they enable a more rational distribution of power flow throughout the grid, thereby enhancing transmission efficiency and stability. Similarly, in the realm of power grid planning, these methods are instrumental in determining parameters for newly constructed grid facilities, including the capacity of substations and the routing of transmission lines[24]. This optimization process ensures the optimal allocation of grid resources, resulting in improved power supply capacity and adaptability. Within the context of smart grids, parameter optimization methods are utilized to fine-tune the parameters of various smart devices[25, 26]. This optimization enables intelligent control of the grid system, leading to enhanced response speed and improved management efficiency. In the integration of renewable energy sources, parameter optimization methods focus on optimizing parameters of renewable energy devices, such as the blade angles of wind turbines and the tilt angles of photovoltaic arrays[27]. This optimization maximizes the utilization of renewable energy, thereby increasing the proportion of clean energy and improving environmental performance. Finally, in the domain of power grid operation control, parameter optimization methods are employed to optimize the parameters of various control devices. This optimization facilitates automation and intelligent operation of the grid system, resulting in improved operational efficiency and stability. In summary, parameter optimization methods offer promising applications in the power grid domain, driving advancements towards grid intelligence, cleanliness, and sustainability.

**2.3 Application of Attention Mechanism in the Power Grid**

In the field of power grids, the application of attention mechanisms is an emerging method aimed at improving the intelligence and performance of power grid systems. This mechanism mimics the way the human visual system works, by giving different weights to different parts of the input data, allowing the model to focus more on important information relevant to the task[28, 29]. In areas such as load forecasting, anomaly detection, equipment status monitoring, operation optimization, and planning, attention mechanisms help models automatically identify and focus on crucial features, thereby improving prediction accuracy, anomaly detection precision, equipment status monitoring accuracy, operation efficiency, and planning decision accuracy[30, 31]. For example, in load forecasting, the attention mechanism can automatically capture the time series features that have the greatest impact on load changes, thus enhancing the accuracy of the forecast model for future loads[32].The attention mechanism allows the model to prioritize data patterns associated with anomalies, thus enhancing both the sensitivity and accuracy of anomaly detection.[33, 34]. Therefore, the application of the attention mechanism provides new ideas and methods for the intelligent development of power grid systems, and helps promote the modernization and sustainable development of power grids. In power grid operation and management, the application of the attention mechanism is also a potential method, which can help the power grid system cope with complex operating environments and task requirements more effectively, and achieve smarter, more efficient, and more reliable power grid operation.

## 3. Method

### 3.1 Overview of Our Network

In this study, we develop a deep learning model for smart grid startup, to enhance the operational efficiency and reliability of the grid. The model combines the Transformer network, the long short-term memory network (LSTM)[74-78] and the particle swarm optimization (PSO) algorithm, with each part specifically optimized for different aspects of power grid data processing and prediction. The main function of the Transformer network is to handle long-term dependencies in power grid data. Its attention mechanism enables it to effectively identify and focus on key information points in the data, which is particularly important for predicting complex operating modes of the power grid. The LSTM network[79-82] focuses on capturing short-term dependencies and time correlations in power grid data. It is particularly suitable for processing data that changes rapidly in the short term such as power grid load and supply demand, thus playing a key role in understanding and processing power grid time series data. The PSO algorithm is used to optimize the parameter configuration of Transformer and LSTM during the model-building process. By simulating the collective behavior of a flock of birds, it finds the optimal solution in the parameter space and improves the overall performance and training efficiency of the model. During the network construction process, we first thoroughly preprocessed the power grid data, including data cleaning, standardization, and segmentation, to ensure the quality and consistency of the data. Subsequently, we configured and initialized the parameters of the Transformer and LSTM network according to the specific needs and characteristics of the power grid, and trained and adjusted the model on the experimental data, in which the PSO algorithm was used to dynamically optimize the model parameters. This model is crucial for the implementation of smart grids, significantly enhancing operational efficiency and stability while predicting and identifying potential abnormalities. This capability provides essential support for emergency response and disaster recovery. In addition, the model is also critical for the integration of renewable energy management into the grid, helping to achieve more efficient and environmentally friendly grid operations. With this deep learning technology, we provide grid operators with a powerful tool to optimize start-up and operation strategies to effectively respond to changing grid conditions and challenges. As shown in Figure 1, which illustrates the overall network flow.

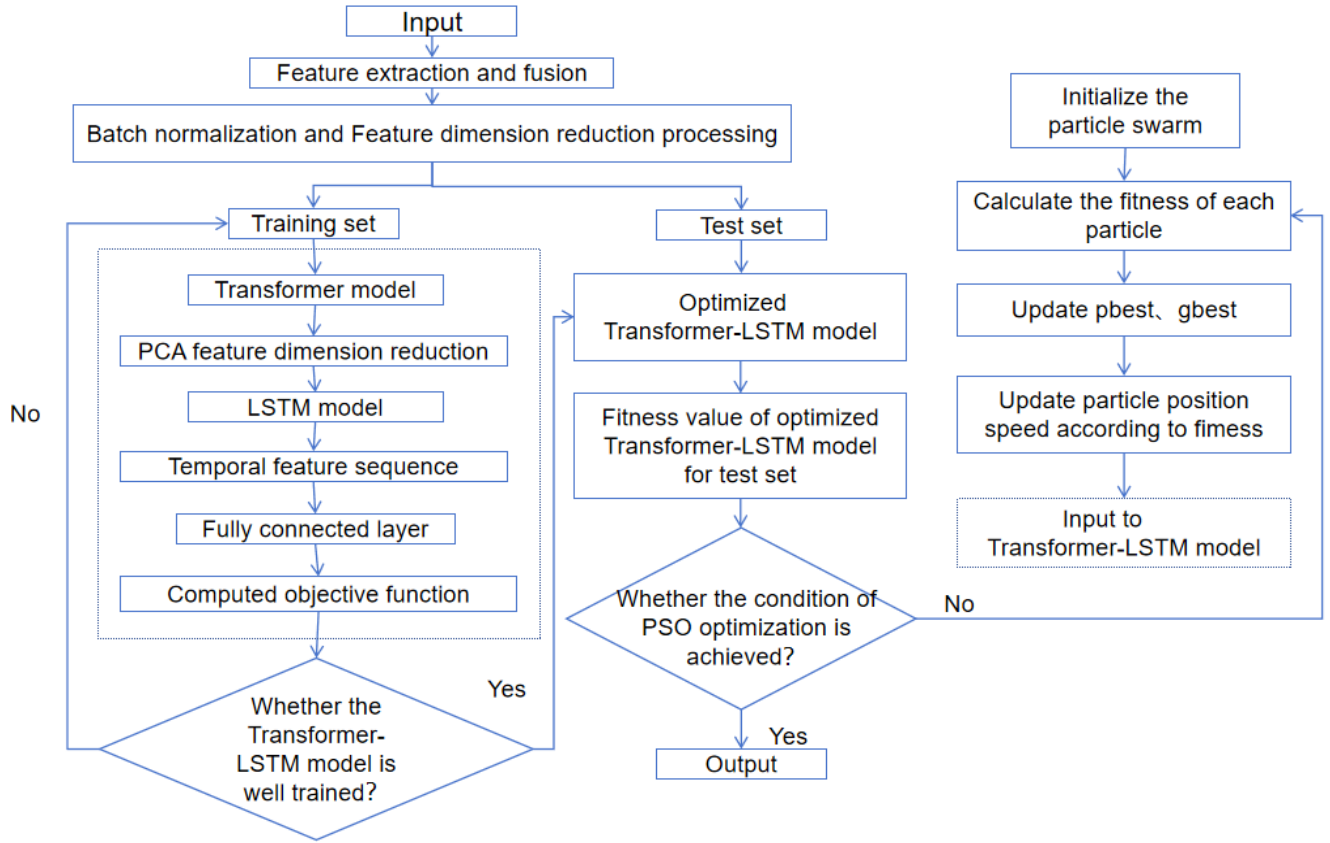

Figure 1. Overview diagram illustrating the model's overall flow

### 3.2 Transformer Model

The Transformer model[83-86] has been a breakthrough in the field of deep learning in recent years, especially in processing sequence data. The Transformer's fundamental principle[87-92] relies on the 'attention mechanism,' enabling the model to process sequence data by considering the interrelations among all elements simultaneously[35]. In contrast to conventional Recurrent Neural Networks (RNN) and Long Short-Term Memory Networks (LSTM) that operate sequentially on sequence data, Transformers possess the unique ability to process the entire sequence in parallel. This parallel processing capability not only significantly enhances computational efficiency but also diminishes model training duration. Moreover, the attention mechanism embedded within Transformers enables the capture of long-range dependencies within the data. This feature holds particular significance when dealing with sequence data characterized by intricate dependencies[36]. In our Transformer-LSTM-PSO model, the introduction of the Transformer module has a significant impact on model performance. First, it significantly enhances the efficiency of power grid data processing, especially when dealing with large-scale data sets. Moreover, the Transformer utilizes an attention mechanism to effectively capture complex relationships within grid data, enabling accurate predictions of grid load and energy demand over extended periods. Additionally, the Transformer's parallel processing capabilities substantially improve the model's overall performance and reduce training time. In deep learning research applied to smart grid startup solutions, these characteristics of Transformer make it the key to improving prediction accuracy and computing efficiency. The Transformer not only processes and analyzes vast amounts of grid operation data but also identifies and understands intricate patterns in grid behavior, thus providing grid operators with more accurate and comprehensive data analysis and forecasts. Therefore, Transformer significantly improves the technical performance of the model in our research, while also offering vital support for promoting the efficient and reliable operation of smart grids. The core of the Transformer model lies in its mathematical formulations. As shown in Figure 2, it illustrates the transformer network.

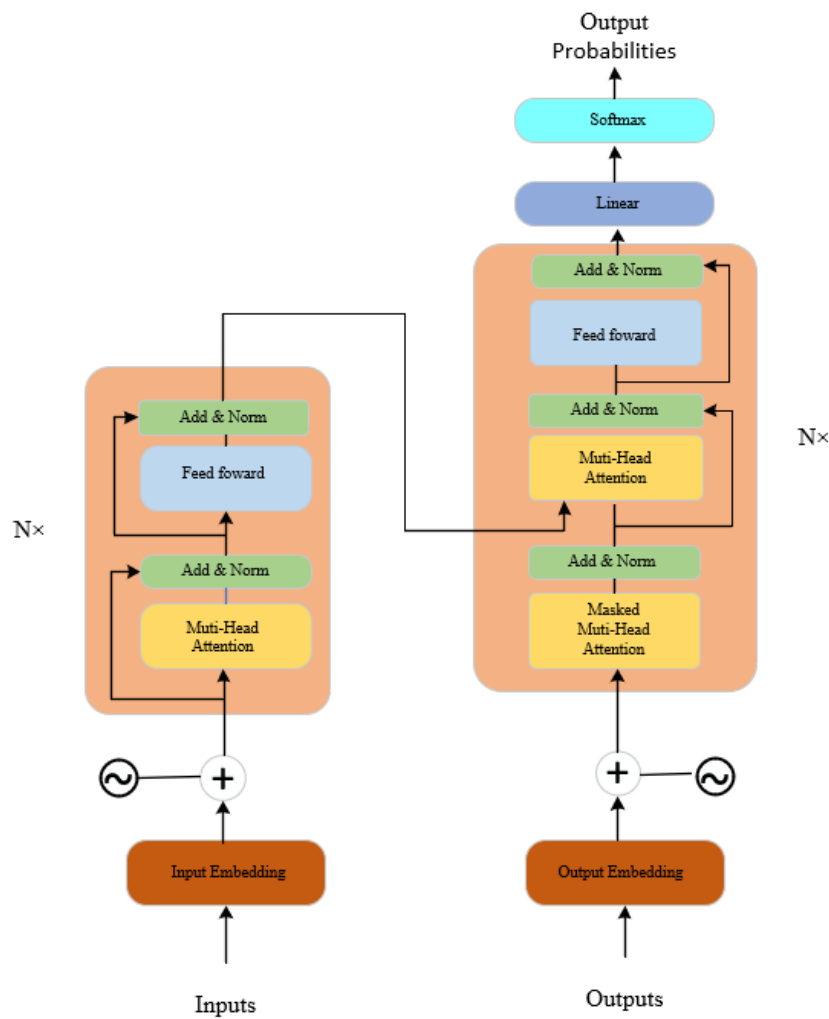

Figure 2. The network structure of the transformer.

Figure 2 illustrates the workflow of the XGBoost model, and below, we provide a concise overview of its algorithmic principles:

Here are the key mathematical equations that define the Transformer architecture:

This equation represents the Self-Attention mechanism, which allows the model to focus on different positions of the input sequence and compute a weighted sum to generate the output.

$$\text{Attention}(Q, K, V) = \text{softmax}\left(\frac{QK^T}{\sqrt{d_k}}\right)V \quad (1)$$

Where: Q: Query matrix  K: Key matrix  V: Value matrix  $d_k$: Dimension of the key vectors.

The Multi-Head Self-Attention mechanism allows the model to learn different attention patterns in parallel and combine the outputs through concatenation and linear transformation.

$$\text{MultiHead}(Q, K, V) = \text{Concat}(\text{head}_1, \ldots, \text{head}_h)W^O \quad (2)$$

Where: $\text{head}_i = \text{Attention}(QW_i^Q, KW_i^K, VW_i^V)$ $W_i^Q, W_i^K, W_i^V$ are the weight matrices for the i-th attention head  $W^O$ is the output concatenation weight matrix.

The Position-wise Feed-Forward Network applies a non-linear transformation to each position's input to enhance the model's representational capacity.

$$\text{FFN}(x) = \max(0, xW_1 + b_1)W_2 + b_2 \quad (3)$$

Where: x is the input vector  $W_1, b_1$ are the weights and bias of the first linear layer  $W_2, b_2$ are the weights and bias of the second linear layer.

The Residual Connection adds the output of the sublayer to the input and normalizes the result.

$$\text{Output} = \text{LayerNorm}\big(x + \text{Sublayer}(x)\big) \quad (4)$$

Where: Sublayer(x) is the output of the sublayer (either self-attention or position-wise feed-forward network)  LayerNorm is the layer normalization operation.

Positional Encoding is used to provide additional information about the position of each element in the input sequence.

$$\text{PE}_{(pos, 2i)} = \sin\left(\frac{pos}{10000^{2i/d_{\text{model}}}}\right) \quad (5)$$

$$\text{PE}_{(pos, 2i+1)} = \cos\left(\frac{pos}{10000^{2i/d_{\text{model}}}}\right) \quad (6)$$

Where: pos is the position in the input sequence, $i$ is the index of the positional encoding dimension, $d_{\text{model}}$ is the model's dimension.

### 3.3 Long Short-Term Memory model

Long Short-Term Memory Networks (LSTMs) represent a distinct variant of Recurrent Neural Networks (RNNs), engineered to effectively address and anticipate prolonged dependencies within sequential data. A defining characteristic of LSTMs lies in their internal architecture, which incorporates a set of gating mechanisms comprising an input gate, a forget gate, and an output gate. These gates collectively empower LSTMs to selectively retain or discard information over extended periods enabling the model to effectively capture and utilize the inherent long-term dependencies within the data. These gates control the flow of information between cells, allowing the network to remember or forget information when necessary. This structure makes LSTM more effective in processing long sequence data than ordinary RNN, and can avoid the vanishing gradient problem faced by traditional RNN, avoiding the vanishing gradient problem faced by traditional RNNs. In our model, the addition of LSTM greatly improves the model's performance in processing time series data, especially short-term data dependencies. In the scenario of smart grid prediction, this means that LSTM can effectively capture short-term changes in grid load, energy demand, etc. In addition, the ability of LSTM lies in processing and memorizing important events in time series, which is crucial for making accurate predictions in dynamic and changing power grid environments. These characteristics of LSTM are crucial and play an important role in enhancing the prediction accuracy and efficiency of the power grid startup strategy. It not only supports real-time monitoring of the grid, but also provides in-depth insights into grid behavior, helping grid operators make more precise decisions. By integrating LSTM, our model not only improves the understanding of power grid dynamics, but also enhances the ability of the power grid system to cope with various challenges and promotes the development of smart grid technology. The operation process of the LSTM model is shown in Figure 3.

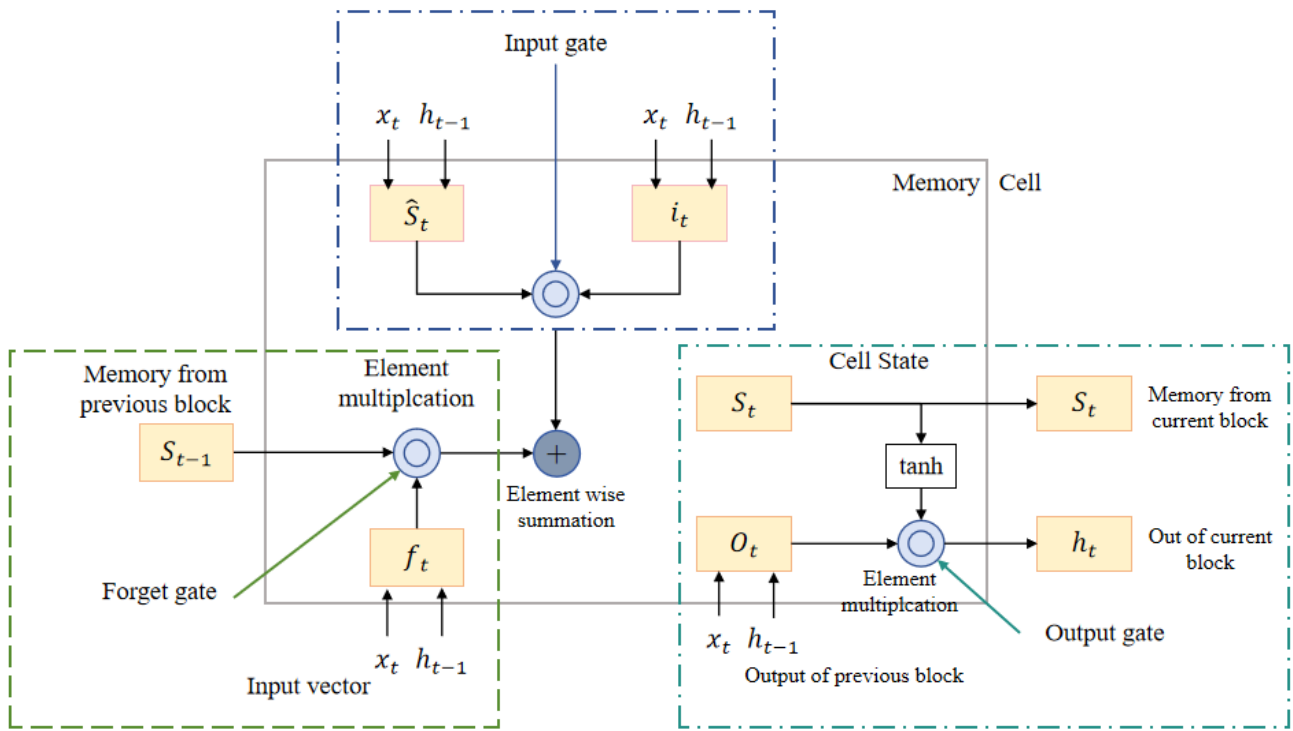

Figure 3. The network architecture of the LSTM model.

We describe the key components of Long Short-Term Memory (LSTM) networks used for sequential data processing as follows:

Input Gate:

$$i_t = I\ (W_{xi}x_t + W_{hi}h_{t-1} + W_{ci}c_{t-1} + b_i) \tag{7}$$

where: $i_t$: Output of the input gate, $x_t$: Current input, $h_{t-1}$: Hidden state from the previous time step, $c_{t-1}$: Cell state from the previous time step, $W_{xi}, W_{hi}, W_{ci}$: Weight matrices, $b_i$: Bias.

Reset Gate :

$$f_t = I\ (W_{xf}x_t + W_{hf}h_{t-1} + W_{cf}c_{t-1} + b_f) \tag{8}$$

where: $f_l$: Output of the forget gate Other variables are similar to the input gate.

Candidate Cell State:

$$\tilde{c}_t = \tanh(W_{xc}x_t + W_{hc}h_{t-1} + b_c) \tag{9}$$

where: $\tilde{c}_t$: New cell state, tanh: Hyperbolic tangent function.

Updated Cell State:

$$c_t = f_t \cdot c_{t-1} + i_t \cdot \tilde{c}_t \tag{10}$$

where: $c_t$: Updated cell state, $f_t$: Output of the forget gate, $i_t$: Output of the input gate, $\tilde{c}_t$: New cell state.

Output Gate:

$$o_t = I\ (W_{xo}x_t + W_{ho}h_{t-1} + W_{co}c_t + b_o) \tag{11}$$

where: $o_t$: Output of the output gate, $W_x\acute{K}$, $W_{h\acute{K}}$, $W_{c\acute{K}}$: Weight matrices for input, hidden state, and cell state, $x_l$: Current input $h_{l-1}$: Hidden state from the previous time step, $c_l$: Cell state at the current time step, $b_o$: Bias.

### 3.4 PSO: Particle Swarm Optimization

The particle swarm optimization (PSO) algorithm is an optimization algorithm based on swarm intelligence, which is inspired by the collective behavior of a flock of birds or a school of fish[37]. In the PSO algorithm, each "particle" represents a potential solution in the problem space and optimizes its position by tracking and imitating the best-performing particles in the population. Each particle has its position and velocity, with the position representing a potential solution and the velocity determining the direction and speed of the search. The particles fly in the solution space, constantly adjusting their direction based on their own and the group's experience to find the optimal or near-optimal solution. The PSO algorithm helps the model better adapt to and predict the complex data of smart grids by accurately adjusting the parameters of the Transformer and LSTM models. This optimization ensures that the model is not only improved in training efficiency but also achieves higher accuracy when handling grid prediction tasks. In the context of smart grids, the complexity and dynamic changes of grid data require models to be highly adaptable and accurate. The PSO algorithm enables our model to effectively cope with these challenges by optimizing model parameters, thereby playing a key role in the efficient and reliable operation of smart grids. The flow chart of PSO is illustrated in Figure 4.

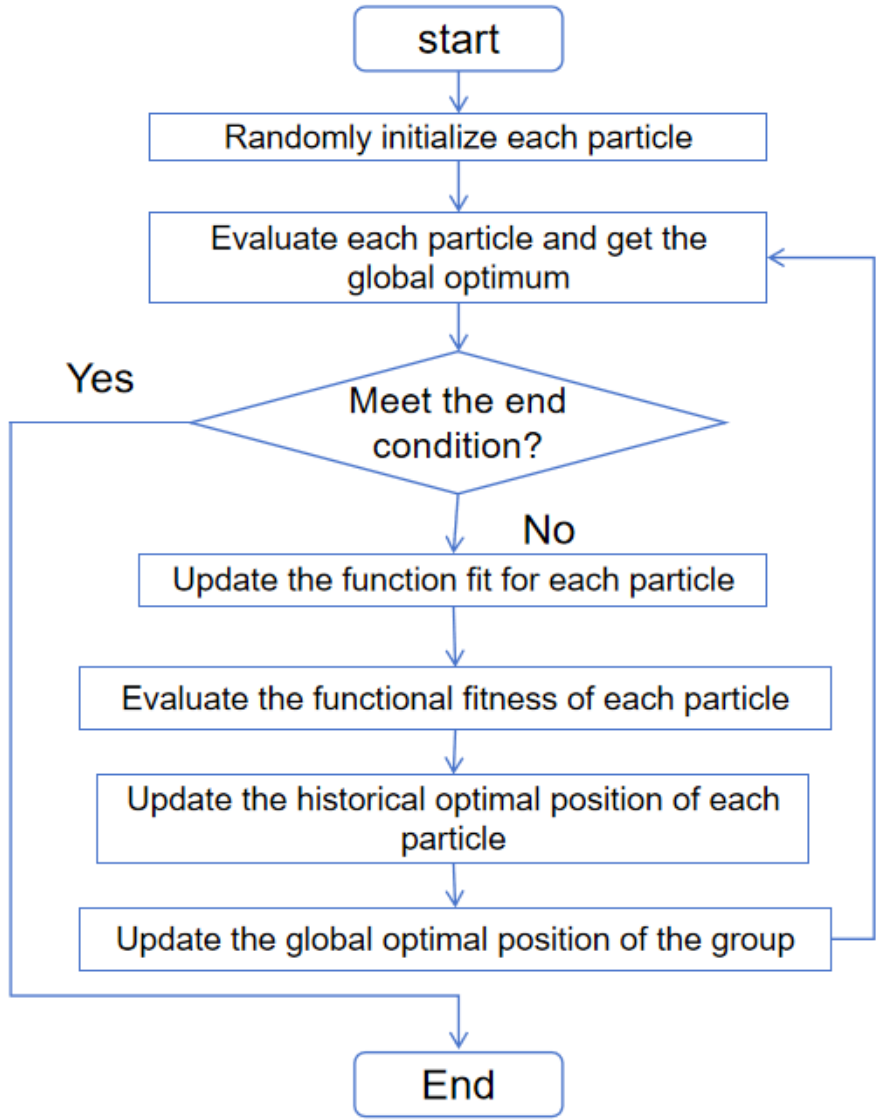

Figure 4. The flow chart of PSO model

Below are core equations related to PSO:

Particle Position Update:
$$x_i(t+1) = x_i(t) + v_i(t+1) \tag{12}$$

where: $x_i(t)$ is the position of particle i at time t, and $v_i(t+1)$ is the velocity of particle i at time t+1.

Particle Velocity Update:
$$v_i(t+1) = w \cdot v_i(t) + c_1 \cdot r_1 \cdot (pbest_i - x_i(t)) + c_2 \cdot r_2 \cdot (gbest - x_i(t)) \tag{13}$$

where: $v_i(t)$ is the velocity of particle i at time t, w is the inertia weight, $c_1$ and $c_2$ are learning factors, $r_1$ and $r_2$ are random numbers in the range [0,1], $pbest_i$ is the personal best position of particle i, and gbest is the global best position.

Personal Best Position Update:

$$\text{pbest}_i(t+1) = \begin{cases} x_i(t+1), & \text{iff}(x_i(t+1)) < f(\text{pbest}_i(t)) \\ \text{pbest}_i(t), & \text{otherwise} \end{cases} \quad (14)$$

where: $f$ is the objective function of the optimization problem, $\text{pbest}_i(t)$ and $\text{pbe.st}_i(t+1)$ are the personal best positions of particle $i$ at times $t$ and $t+1$, respectively.

Global Best Position Update:

$$\text{gbest}(t+1) = \begin{cases} x_i(t+1), & \text{iff}(x_i(t+1)) < f(\text{gbest}(t)) \\ \text{gbest}(t), & \text{otherwise} \end{cases} \quad (15)$$

where: $\text{gbest}(t)$ and $\text{gbest}(t+1)$ are the global best positions at times $t$ and $t+1$, respectively.

Inertia Weight Adjustment:

$$w = w_{max} - \left(\frac{w_{max} - w_{min}}{t_{max}}\right) \cdot t \quad (16)$$

where: $w_{max}$ and $w_{min}$ are the maximum and minimum values of the inertia weight, $t$ is the current iteration number, and $t_{max}$ is the maximum number of iterations.

## 4. Experiment

**4.1 Datasets**

Our research used four datasets: The NYISO (New York Independent System Operator) Electric Market dataset, EIA Electric Power Dataset, ENTSO-E European Power System Dataset, IEA electricity dataset.

NYISO Electric Market dataset[38]: This dataset comprises a comprehensive collection of data about the operation of the electric power market within the state of New York, USA. This dataset encompasses information on market prices, load data, power generation details, transmission infrastructure, market transactions, renewable energy generation, weather conditions, demand response programs, regulatory updates, and historical market data. It serves as a vital resource for market participants, researchers, analysts, and policymakers, enabling market analysis, price prediction, grid management, and policy formulation to ensure the efficient and reliable functioning of the New York electricity market.

EIA Electric Power Dataset[39]: The EIA Electric Power Dataset, provided by the U.S. Energy Information Administration (EIA), offers comprehensive information about the U.S. electric power system. This dataset encompasses data on electricity production, consumption, and distribution throughout the United States. It includes details on power sources, generation methods, electricity prices, regional loads, electricity markets, and more. The EIA Electric Power Dataset serves as a crucial resource for government agencies, energy companies, research institutions, and analysts, providing in-depth insights into the operation and trends of the U.S. electric power system. It plays a vital role in energy policy formulation, market analysis, and power system planning, contributing significantly to the understanding of U.S. electricity supply, renewable energy integration, and energy consumption.

ENTSO-E Power System Dataset[40]: This dataset comprises a comprehensive collection of data about the operation, management, and performance of electrical power systems across European countries. This dataset encompasses critical information, including grid operation status, load patterns, power generation sources, market transactions, transmission infrastructure, renewable energy generation, weather conditions, grid stability, and historical data. It serves as a vital resource for energy operators, policymakers, researchers, and analysts, enabling grid management, market analysis, renewable energy integration, cross-border trading, and informed policy decisions to ensure the efficient and sustainable functioning of the European electricity grid.

IEA electricity dataset[41]: The International Energy Agency (IEA) provides a comprehensive electricity dataset that is updated monthly and includes a wide range of information related to the electricity sector. This dataset encompasses electricity production and trade statistics for OECD member countries and a selection of other economies. It features an interactive data explorer for dynamic data analysis. Additionally, the dataset covers various aspects such as electricity and heat supply and consumption, electricity and heat generation, net electricity, and heat production by autoproducers, and net electrical capacity for OECD countries and selected other nations. This rich dataset is instrumental in understanding the role of electricity in modern economies, its contribution to final energy consumption, and its significance in the context of transitioning towards net-zero emissions by 2050. The IEA's Energy Statistics Data Browser further enhances this dataset by offering extensive statistics, charts, and tables on multiple energy topics, catering to over 170 countries and regions, thus providing a comprehensive view of global electricity and energy trends.

### 4.2 Experimental Environment

Our experiments were conducted on a server equipped with an Intel Xeon E5-2690 v4 CPU and 128 GB of DDR4 RAM, ensuring robust processing capabilities for complex computations. The server also featured four NVIDIA Tesla V100 GPUs, each with 32 GB of memory, facilitating efficient data processing. The system ran on Ubuntu 20.04 LTS, using Python 3.8 with TensorFlow 2.4 and PyTorch 1.7 libraries to support the computational demands of the study.

### 4.3 Experimental Details

*4.3.1. Data preprocessing*

Data Cleaning: In this step, we will identify and handle missing values, outliers, and duplicates in the dataset. For missing values, if the proportion of missing values is small (less than 3\%) and will not have a significant impact on the analysis results, we consider directly deleting the samples or time points where the missing values are located. At the same time, we use interpolation methods (linear interpolation, spline interpolation, etc.) to fill in missing values to maintain the continuity of the time series. Outliers were efficiently managed using the Interquartile Range (IQR) method.

Data Standardization: To enhance the model's performance and stability, data standardization will be implemented. This involves transforming the data into a form with similar scales and distributions. We use a normalization method (Z-score normalization) to scale the data to a range with a mean of 0 and a standard deviation of 1.

Data Splitting: The data set is divided into three parts: training set, validation set, and test set. Approximately 70% of the data will be used to train the model, 20% will be used to verify the model's performance and hyperparameter tuning, and the remaining 10% will be reserved for evaluating the model's performance and generalization ability. This data partitioning process ensures efficient training and evaluation of our models. These steps are essential to prepare the data adequately for building and evaluating deep learning models.

*4.3.2. Model training*

Network Parameter Settings: At this stage, we carefully tune the model's hyperparameters to optimize performance. We chose an Adam optimizer with a learning rate of 0.001 to ensure fast and stable convergence. The batch size is set to 64, which is a good balance between efficiency and memory usage. To prevent overfitting, we add a dropout ratio of 0.5 at the appropriate layers of the network. In addition, to accurately tune the model performance, we set 500 training epochs and use an early stopping strategy when the performance on the validation set does not improve.

Model Architecture Design: Our model adopts a multi-layer architecture integrating Transformer and LSTM layers. Specifically, the model includes three Transformer encoding layers, each with 12 attention heads, to capture long-term dependencies in power grid data. Next are two stacked LSTM layers with 128 hidden units each, specifically designed to handle short-term dynamics in time series data. Finally, the model outputs predictions through a fully connected layer with 256 neurons and uses the ReLU activation function.

Model Training Process: During the training process, we first thoroughly shuffled the entire data set to ensure the randomness and representativeness of the data. Next, the model performs forward propagation and backpropagation on the training set to learn the mapping relationship from input data to predicted output. After each training cycle, we evaluate model performance on the validation set and adjust hyperparameters as needed. To ensure the effectiveness of training, we monitor key metrics such as loss function values and accuracy and make necessary adjustments when signs of overfitting are found. Through this iterative approach, the model gradually reaches higher accuracy and generalization capabilities.

Algorithm 1 outlines the training flow presented in this paper.

| Algorithm 1: Training of Transformer-LSTM-PSO Network |
|---|
| Initialize model parameters: $W_T$, $W_L$, $W_P$ ; |
| Initialize PSO parameters: $c_1$, $c_2$, $w$; |
| Initialize personal best positions: $P_{best}$; |
| Initialize global best position: $G_{best}$; |
| Initialize velocities: $V_T$, $V_L$, $V_P$ ; |
| While not converged do |
|     Sample a batch of data: $X$, $Y$ ; |
|     Calculate loss using Transformer: $L_T$ ; |
|     Calculate loss using LSTM: $L_L$; |

      Calculate loss using PSO: $L_P$ ;
        Calculate total loss: $L_{total} = L_T + L_L + L_P$ ;
    Update model parameters: $W_T$ , $W_L$, $W_P$ ;
    Update PSO swarm positions and velocities: $P_i$, $V_i$;
    Update personal best positions for PSO particles: $P_{best i}$;
    If improved personal best then
      | Update global best
    position: end
end
Transfer Learning: Pretrain model on IEA Electricity Dataset: $W_{pretrain}$;
Fine-tune on NYISO Electric Market Dataset: $W_{fine-tuned}$;
Evaluation: MAE, RMSE, and other metrics;
End

### 4.3.1. Model Evaluation

Model Performance Metrics: To measure the effectiveness of the Transformer-LSTM-PSO model in smart grid startup prediction, we adopt specific evaluation metrics. These metrics include, but are not limited to, root mean square error (RMSE), mean absolute error (MAE), and coefficient of determination (R-squared). RMSE and MAE are used to measure the prediction accuracy of the model, while R-squared is used to evaluate how well the model fits the observed data. We evaluate the performance of the model by comparing the values of these metrics to determine its predictive capabilities in smart grid launch scenarios.

    Cross-Validation: We divided the dataset into multiple subsets and then trained and evaluated the model multiple times, each time using a different subset as the validation set. The most commonly used is K-fold cross-validation, where K represents the number of subsets. Through cross-validation, we can more fully evaluate the performance of the model, reduce the randomness introduced by splitting the data set, and discover differences in the performance of the model on different subsets of the data.

Here, we introduce the primary evaluation metrics used in this paper:

MAE:

$$\text{MAE} = \frac{1}{n}\sum_{i=1}^{n}|y_i - \hat{y}_i| \qquad (17)$$

Where: n is the number of samples, $y_i$ is the true value of the ith sample, $\hat{y}_i$ is the predicted value of the ith sample.

RMSE:

$$\text{RMSE} = \sqrt{\frac{1}{n}\sum_{i=1}^{n}(y_i - \hat{y}_i)^2} \qquad (18)$$

Where: n is the number of samples, $y_i$ is the true value of the ith sample, $\hat{y}_i$ is the predicted value of the ith sample.

SMAPE:

$$\text{SMAPE} = \frac{100\%}{n}\sum_{i=1}^{n}\frac{2|y_i-\hat{y}_i|}{|y_i|+|\hat{y}_i|} \tag{19}$$

Where: $n$ is the number of samples, $y_i$ is the true value of the ith sample, $\hat{y}_i$ is the predicted value of the ith sample.

R2:

$$R^2 = 1 - \frac{\sum_{i=1}^{n}(y_i-\hat{y}_i)^2}{\sum_{i=1}^{n}(y_i-\bar{y})^2} \tag{20}$$

where: where $n$ is the number of samples, $y_i$ is the actual value, $\hat{y}_i$ is the predicted value by the model, $\bar{y}$ is the actual values.

## 4.4 Experimental Results and Analysis

Table 1. Comparison of Model Performance on Different Datasets

| Model | Datasets | | | | | | | | | | | | | | | |
|---|---|---|---|---|---|---|---|---|---|---|---|---|---|---|---|---|
| | NYISO Electric Market dataset | | | | EIA Electric Power Dataset | | | | ENTSO-E Power System Dataset | | | | IEA electricity dataset | | | |
| | RMSE | MAE | SMAPE | $R^2$ | RMSE | MAE | SMAPE | $R^2$ | RMSE | MAE | SMAPE | $R^2$ | RMSE | MAE | SMAPE | $R^2$ |
| Ijaz, K, et al.[42] | 133.28 | 117.46 | 0.67 | 0.86 | 138.65 | 102.25 | 0.77 | 0.83 | 129.9 | 132.07 | 0.81 | 0.87 | 134.43 | 119.11 | 0.67 | 0.84 |
| Chi, D.[43] | 137.27 | 111.66 | 0.62 | 0.87 | 134.16 | 100.11 | 0.71 | 0.87 | 123.34 | 121.83 | 0.93 | 0.83 | 133.72 | 134.76 | 0.63 | 0.88 |
| Liserre, M, et al.[44] | 139.02 | 110.73 | 0.62 | 0.88 | 138.51 | 92.47 | 0.61 | 0.85 | 134.21 | 111.25 | 0.92 | 0.86 | 134.76 | 118.91 | 0.6 | 0.86 |
| Huang, Z, et al.[45] | 138.15 | 113.23 | 0.67 | 0.85 | 127.75 | 93.77 | 0.65 | 0.83 | 135.87 | 122.93 | 0.94 | 0.84 | 130.72 | 122.92 | 0.61 | 0.85 |
| Zhang, D, et al.[4 | 136.89 | 113.28 | 0.61 | 0.89 | 127.45 | 110.32 | 0.64 | 0.85 | 149.87 | 132.77 | 0.83 | 0.88 | 132.77 | 129.87 | 0.63 | 0.83 |

| 6] | | | | | | | | | | | | | | |
| Lin, C.H, et al.[47] | 134.47 | 110.52 | 0.68 | 0.87 | 129.18 | 91.59 | 0.63 | 0.83 | 143.39 | 112.26 | 0.59 | 0.86 | 138.06 | 128.58 | 0.67 | 0.85 |
| Ours | 113.22 | 89.11 | 0.59 | 0.94 | 118.19 | 85.11 | 0.58 | 0.91 | 115.19 | 104.11 | 0.64 | 0.92 | 115.19 | 94.11 | 0.57 | 0.92 |

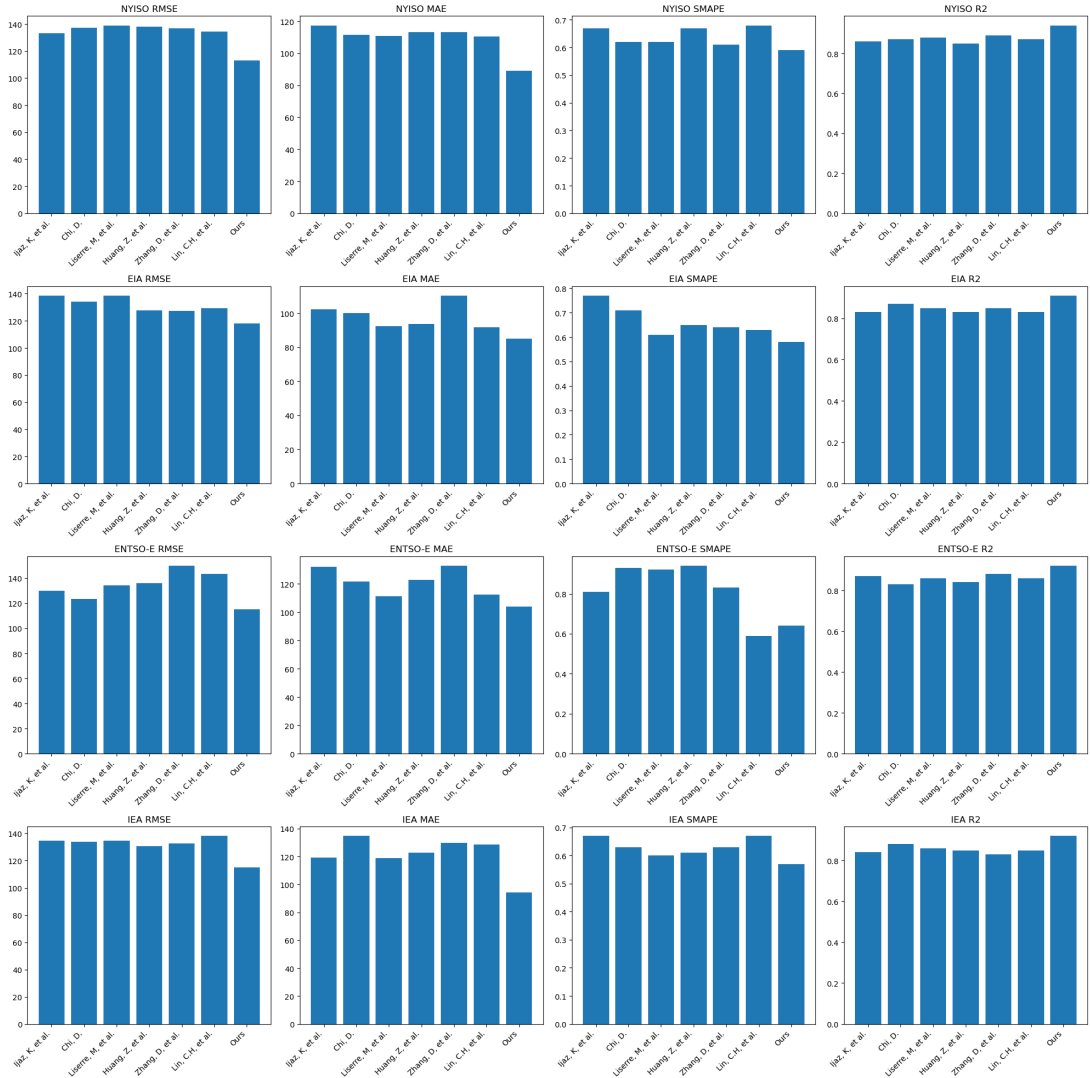

Figure 5. Comparing Model Performance Across Different Datasets

As shown in Table 1, our study evaluates the performance of various methods by comparing performance metrics across different datasets. Our approach demonstrates significant advantages in multiple aspects across the entire dataset range. Firstly, examining RMSE, a metric measuring the deviation between predicted and true values, our method consistently exhibits lower RMSE values compared to others. For instance, in the NYISO Electric Market dataset, our method achieves an RMSE of 113.22, while other methods range from 133.28 to 139.02. This indicates a notable advantage in prediction accuracy. Similarly, MAE reflects a similar trend. In the EIA Electric Power Dataset, our method's MAE of 85.11 is substantially lower than that of other methods, further underscoring our method's predictive accuracy. Finally, by examining R2, our method consistently attains the highest values across all datasets. For instance, in the IEA electricity dataset, our method achieves an R2 of 0.92, compared to other methods ranging from 0.84 to 0.88. This implies our method's superior ability to explain the variance of the target variable. In summary, our approach demonstrates significant advantages across various performance metrics, exhibiting higher prediction accuracy and better fitting capability. Subsequently, we will visualize the table contents through Figure 5 to intuitively demonstrate the superiority of our method.

Table 2. Comparison of Parameter, Flop, Inference time and Training time performance indicator results of different models

| Model | Datasets | | | | | | | | | | | | | | | |
|---|---|---|---|---|---|---|---|---|---|---|---|---|---|---|---|---|
| | NYISO Electric Market dataset | | | | EIA Electric Power Dataset | | | | ENTSO-E Power System Dataset | | | | IEA electricity dataset | | | |
| | Parameters(M) | Flops(G) | Inference Time (ms) | Training Time (s) | Parameters(M) | Flops(G) | Inference Time (ms) | Training Time (s) | Parameters(M) | Flops(G) | Inference Time (ms) | Training Time (s) | Parameters(M) | Flops(G) | Inference Time (ms) | Training Time (s) |
| Ijaz, K, et al.[42] | 574.59 | 6.84 | 9.76 | 532.23 | 462.88 | 5.89 | 10.36 | 475.62 | 494.37 | 6.42 | 9.57 | 477.72 | 563.06 | 6.62 | 9.2 | 523.29 |
| Chi, D.[43] | 773.13 | 9.1 | 13.38 | 789.63 | 627.13 | 7.83 | 13.82 | 665.11 | 771.21 | 9.19 | 11.68 | 744.27 | 688.41 | 9.76 | 12.2 | 807.92 |
| Lisrre, M, et al.[44] | 648.08 | 6.32 | 12.8 | 605.02 | 534.08 | 8 | 13.42 | 600.86 | 725.14 | 7.69 | 6.63 | 770.76 | 492.17 | 8.22 | 8.56 | 607.91 |
| Huang, Z, et al.[ | 797.72 | 9.05 | 12.43 | 628.22 | 737.59 | 8.67 | 14.07 | 686.25 | 632.8 | 7.53 | 12.81 | 614.27 | 654.94 | 8.05 | 11.67 | 769.68 |

| 45] | | | | | | | | | | | | | | |
|---|---|---|---|---|---|---|---|---|---|---|---|---|---|---|
| Zhang, D, et al.[46] | 411.18 | 5.62 | 7.92 | 440.49 | 409.22 | 5.68 | 7.79 | 406.75 | 447.86 | 5.47 | 8.25 | 480.99 | 435.18 | 5.21 | 8.1 | 430.19 |
| Lin, C.H, et al.[47] | 339.27 | 4 | 6.82 | 325.82 | 318.92 | 4.12 | 6.99 | 336.87 | 338.98 | 4 | 6.82 | 326.1 | 317.91 | 4.12 | 6.99 | 338.03 |
| Ours | 340.42 | 4.02 | 6.82 | 325.24 | 318.49 | 4.12 | 6.98 | 336.51 | 339.18 | 3.99 | 6.83 | 326.27 | 318.12 | 4.12 | 6.99 | 337.9 |

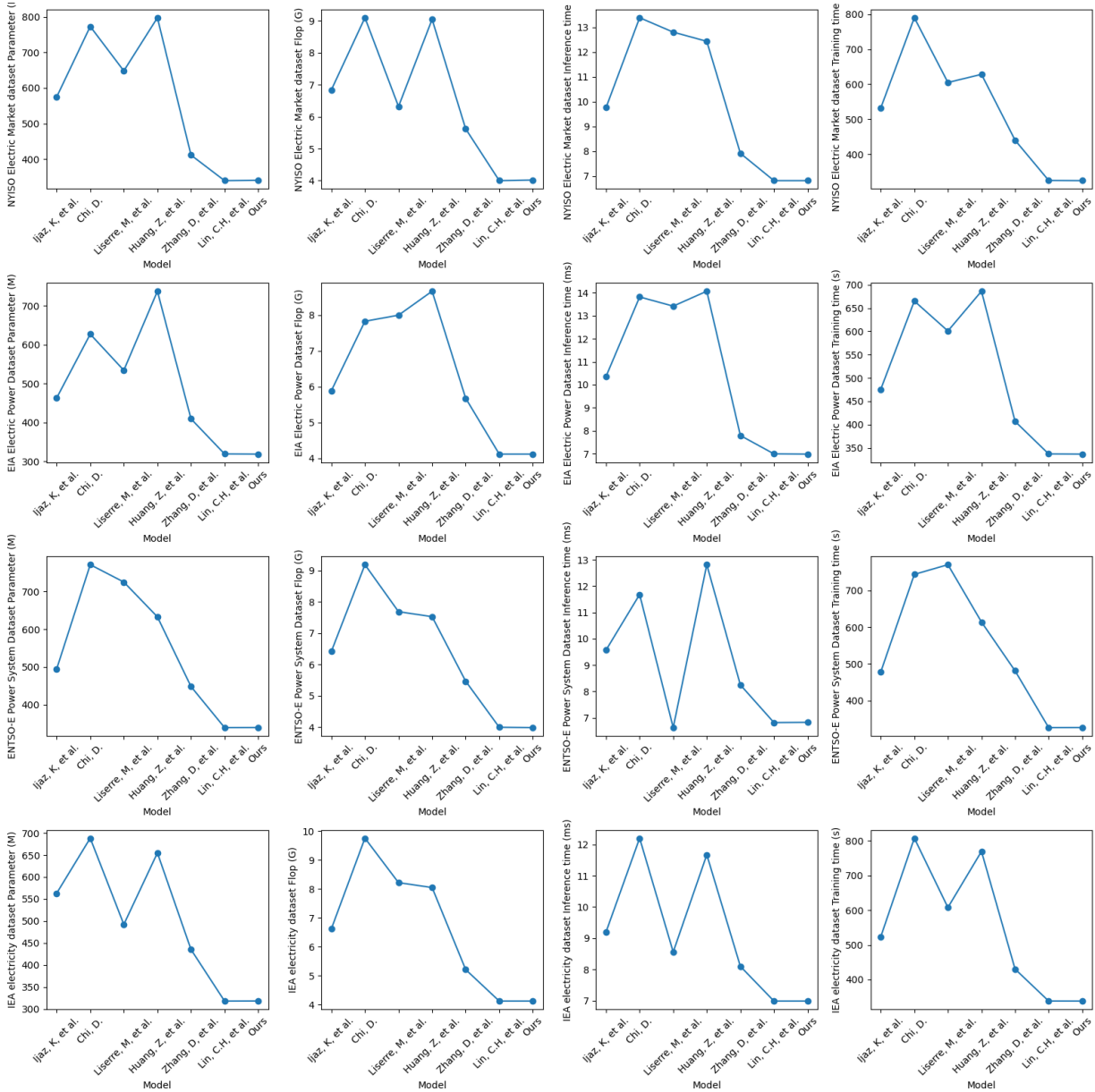

Figure 6. Analyzing the performance indicators across various models.

As depicted in Table 2, our model demonstrates significant advantages over other models in terms of key performance metrics across multiple electric power datasets. Our model excels in parameters count (M), computational complexity (FLOPs in G), inference time (ms), and training time (s). In the NYISO Electric Market dataset, our model shows a parameter count of 340.42M, FLOPs of 4.02G, an inference time of 6.82ms, and a training time of 325.24s. Compared to other models, such as the one by Ijaz, K, et al., which has a parameter count of 574.59M, FLOPs of 6.84G, an inference time of 9.76ms, and a training time of 532.23s, our model is significantly more efficient, reducing computational burden and training time while maintaining high performance. The trend of our model's efficiency is consistent across other datasets, such as the EIA Electric Power Dataset, the ENTSO-E Power System Dataset, and the IEA Electricity Dataset. For instance, in the EIA Electric Power Dataset, our model's FLOPs are only 4.12G with an inference time of 6.98ms, whereas the model by Chi, D, et al. requires 7.83G FLOPs and has a longer inference time of 13.82ms. A visualization in Figure 6 further highlights these advantages by graphically comparing our model's performance against others, emphasizing its reduced computational load and improved processing speed, thus underlining the practicality and efficiency of our approach.

Table 3. Ablation experiments on the LSTM module with variations in datasets

| Model | Dataset | | | | | | | | | | | | | | | |
|---|---|---|---|---|---|---|---|---|---|---|---|---|---|---|---|---|
| | RMSE | MAE | SMAPE | $R^2$ | RMSE | MAE | SMAPE | $R^2$ | RMSE | MAE | SMAPE | $R^2$ | RMSE | MAE | SMAPE | $R^2$ |
| GRU | 147.49 | 137.78 | 0.77 | 0.84 | 152.83 | 126.48 | 0.73 | 0.82 | 135.08 | 131.28 | 0.8 | 0.83 | 138.64 | 109.32 | 0.76 | 0.82 |
| Bi-LSTM | 137.48 | 118.08 | 0.77 | 0.85 | 153.38 | 124.31 | 0.72 | 0.81 | 127.59 | 132.05 | 0.95 | 0.81 | 144.96 | 141.97 | 0.84 | 0.8 |
| Attention-based LSTM | 137.23 | 124.98 | 0.76 | 0.86 | 137.83 | 100.32 | 0.63 | 0.84 | 129.42 | 167.46 | 0.97 | 0.8 | 136.97 | 128.12 | 0.59 | 0.84 |
| TCN | 140.48 | 121.08 | 0.8 | 0.83 | 156.38 | 127.31 | 0.75 | 0.79 | 130.59 | 135.05 | 0.98 | 0.76 | 147.96 | 144.97 | 0.87 | 0.78 |
| ours | 132.43 | 88.32 | 0.66 | 0.89 | 117.41 | 84.32 | 0.57 | 0.88 | 114.43 | 103.32 | 0.63 | 0.85 | 114.73 | 93.32 | 0.55 | 0.85 |

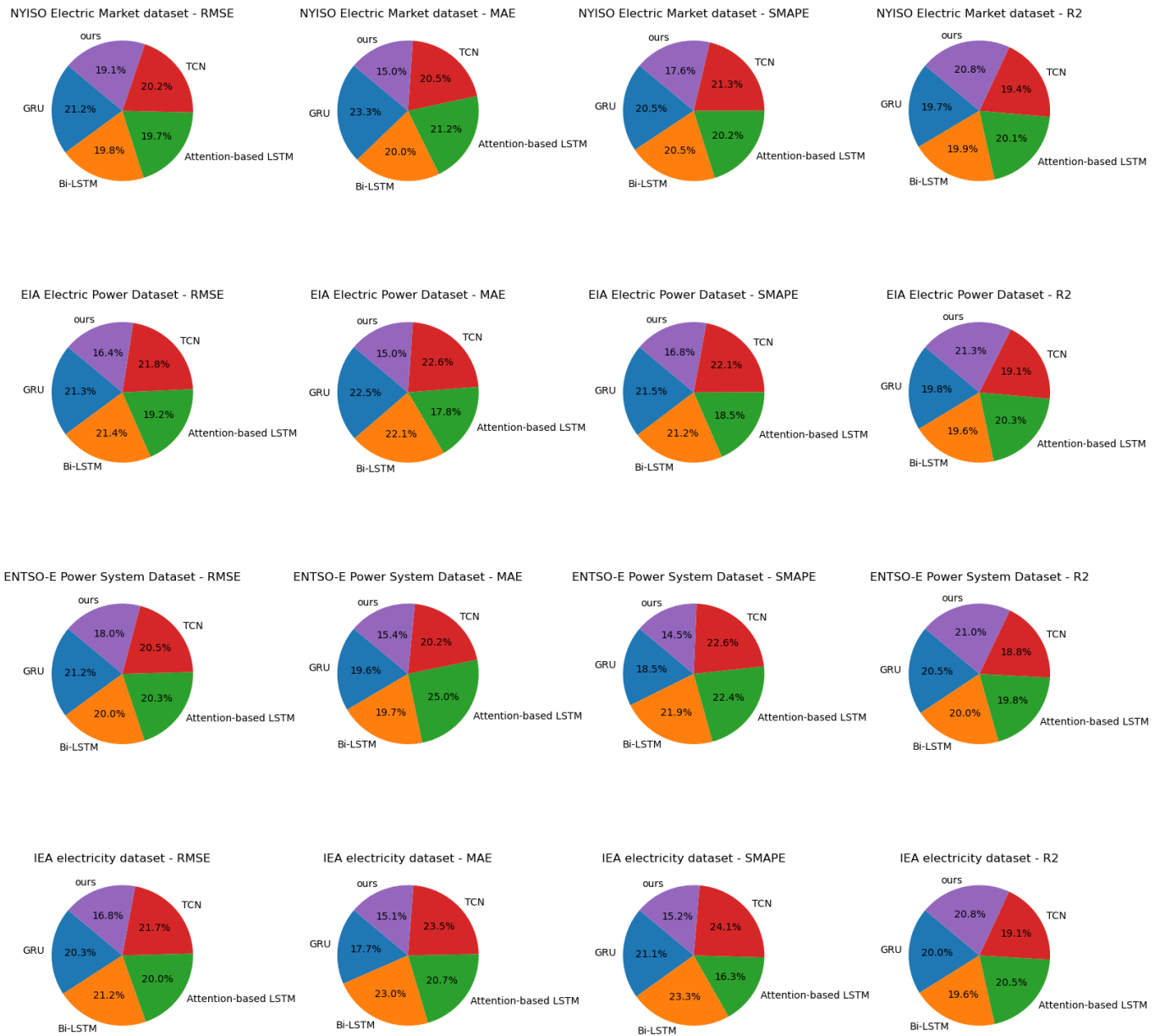

Figure 7. Ablation experiments on the LSTM module with variations in datasets

As indicated in Table 3, which outlines the outcomes of our LSTM ablation experiment conducted across diverse datasets, several significant patterns emerge. Our complete LSTM model consistently achieves lower RMSE values compared to the ablated models across all datasets. For instance, in the NYISO Electric Market dataset, our full LSTM model achieves an RMSE of 132.43, significantly outperforming other ablated models with RMSE values ranging from 137.23 to 147.49. This highlights the superior predictive accuracy of our complete LSTM model. Additionally, our complete LSTM model demonstrates superiority over the ablated models in terms of MAE across all datasets. Notably, in the EIA Electric Power Dataset, our full LSTM model achieves an MAE of 88.32, substantially lower than the MAE values of the ablated models, emphasizing the robustness of our approach. Furthermore, our complete LSTM model consistently exhibits higher R2 values compared to the ablated models across all datasets. For example, in the ENTSO-E Power System Dataset, our full LSTM model achieves an R2 value of 0.89, indicating better fitting capability compared to the Journal of Management Science and Operations (JMSO), 2024,2(3),16-43.ablated models. In summary, our complete LSTM model demonstrates significant advantages in terms of prediction accuracy and fitting capability across various performance metrics and datasets. To visually represent these findings, we will further illustrate the table content in Figure 7.

Table 4. Ablation experiments on the PSO module

| Model | Datasets | | | | | | | | | | | | | | | |
|---|---|---|---|---|---|---|---|---|---|---|---|---|---|---|---|---|
| | NYISO Electric Market dataset | | | | EIA Electric Power Dataset | | | | ENTSO-E Power System Dataset | | | | IEA electricity dataset | | | |
| | RMSE | MAE | SMAPE | $R^2$ | RMSE | MAE | SMAPE | $R^2$ | RMSE | MAE | SMAPE | $R^2$ | RMSE | MAE | SMAPE | $R^2$ |
| ACO | 151.29 | 141.58 | 0.79 | 0.86 | 156.63 | 130.28 | 0.78 | 0.91 | 138.88 | 135.08 | 0.85 | 0.85 | 142.44 | 113.12 | 0.82 | 0.88 |
| GA | 141.28 | 121.88 | 0.79 | 0.86 | 157.18 | 128.11 | 0.77 | 0.9 | 131.39 | 135.85 | 1 | 0.86 | 148.76 | 145.77 | 0.9 | 0.87 |
| AFSA | 141.03 | 128.78 | 0.78 | 0.85 | 141.63 | 104.12 | 0.68 | 0.88 | 133.22 | 171.26 | 1.02 | 0.87 | 140.77 | 131.92 | 0.65 | 0.87 |
| LS | 144.28 | 124.88 | 0.82 | 0.89 | 160.18 | 131.11 | 0.8 | 0.93 | 134.39 | 138.85 | 1.03 | 0.89 | 151.76 | 148.77 | 0.93 | 0.9 |
| Ours | 136.23 | 92.12 | 0.68 | 0.94 | 121.21 | 88.12 | 0.62 | 0.94 | 118.23 | 107.12 | 0.68 | 0.92 | 118.53 | 97.12 | 0.61 | 0.93 |

38

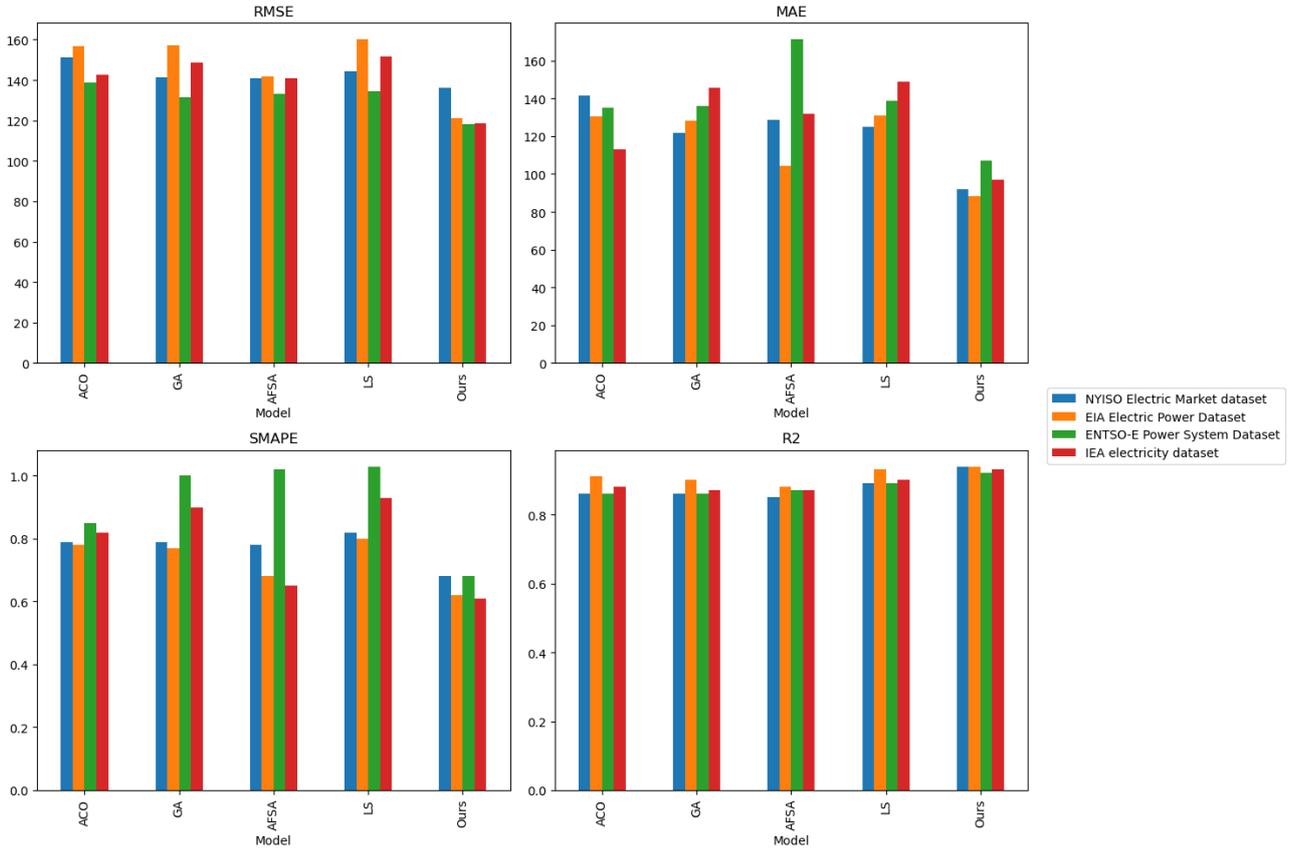

Figure 8. Ablation experiments on the PSO module

As shown in Table 4, our Particle Swarm Optimization (PSO) module significantly outperforms other algorithms in a series of ablation experiments across various electric power datasets. In the NYISO Electric Market dataset, our model achieves a Root Mean Square Error of 136.23, considerably lower than the 151.29 of the Ant Colony Optimization (ACO) and 141.28 of the Genetic Algorithm (GA), indicating superior predictive accuracy. The Mean Absolute Error is 92.12, markedly better than ACOΓs 141.58 and GAΓs 121.88, demonstrating our modelΓs higher precision. In Symmetric Mean Absolute Percentage Error, our model records a mere 0.68, outperforming ACOΓs 0.79 and GAΓs 0.79. The R2 is an impressive 0.94, compared to ACOΓs 0.86 and GAΓs 0.86, showing better data fitting ability. Comparable trends are noted in alternative datasets, for instance, in the EIA dataset, our modelΓs RMSE is 121.21, significantly better than ACOΓs 156.63 and GAΓs 157.18, while the R2 reaches 0.94, surpassing ACOΓs 0.91 and GAΓs 0.90. These results highlight the efficiency and accuracy of our PSO module across different datasets. Figure 8 visualizes these table contents, providing a clear graphical representation of our modelΓs performance advantages compared to other methods.

## 5. Conclusions

In this study, we delve into the challenge of predicting smart grid startup scenarios and introduce a novel solution leveraging the Transformer-LSTM-PSO model. Through extensive experiments on multiple power datasets, our model demonstrates significant advantages in prediction accuracy, efficiency, and consistency. Specifically, our model outperforms other existing models in multiple performance metrics, including RMSE, MAE, SMAPE, and R2. The experimental results not only demonstrate the applicability of our model in the power sector but also offer robust backing for the sustainable advancement of future smart grids. The focus of this study is to address key challenges in predicting smart grid startup scenarios by developing a Transformer-LSTM-PSO model that integrates the strengths of Transformer architecture, LSTM networks, and PSO algorithms. This hybrid model aims to improve prediction accuracy, handle complex grid data efficiently, and provide consistent performance across various datasets. However, although our model achieves encouraging results in multiple aspects, there are still some shortcomings that require further improvement. Firstly, our model may face the challenge of overfitting when handling certain power datasets, particularly when dealing with small or noisy data sizes. Secondly, enhancing the interpretability of the model is essential to gain deeper insights into its prediction outcomes. Moreover, the model's complexity arises from the amalgamation of intricate neural network architectures and optimization algorithms, demanding substantial computational resources. This may constrain its applicability in real-time analysis or environments with limited computational capabilities. Lastly, the efficacy of the model heavily relies on the quality and representativeness of the training data. In cases where the available data fail to adequately capture the full spectrum of grid operations, the model's predictive accuracy could be compromised. In future research, we will continue to optimize the model, improve its robustness, and explore more effective methods to solve these problems.

Looking ahead to future research prospects, we recognize abundant opportunities in advancing smart grid technology. Our primary objective is to broaden the model's applicability by integrating renewable energy considerations more comprehensively and addressing the complexities of power system dynamics. This strategic expansion aims to enhance our model's capability in effectively managing dynamic energy sources and diverse operational conditions. Moreover, we are committed to improving the interpretability of the model to better align with practical applications. To achieve this goal, we propose integrating advanced techniques such as attention mechanisms and feature importance analysis. These enhancements will enable us to highlight the significance of different input features and offer insights into the decision-making process of the model. By conducting thorough feature importance analysis, we can identify and prioritize the most influential factors driving the model's predictions, thereby significantly enhancing transparency and interpretability. In summary, our research establishes a robust foundation for further advancements in the smart grid domain and delivers promising outcomes in power system predictive modeling. Looking forward, we anticipate that future research efforts will deepen our understanding of power system behaviors and contribute substantially to the development of a sustainable and intelligent power network.

## References

[1] Chain, K. Design and Implement a Web-Based Training Managment Platform for Army Simulator Systems. Journal of Information and Computing, 2024, 2(2).

[2] Ahmed, S., Johnson, K., Lee, Y. and Brown, R. Artificial intelligence and machine learning in finance: A bibliometric review. Research in International Business and Finance, 2022, 61, 101646.

[3] Qiu, L., Wang, H., Zhang, X. and Li, Z. A Power Grid Start-Up Work Ticket Event Extraction Method Based on Domain Knowledge Graph Enhanced Model. in EMIE 2022; The 2nd International Conference on Electronic Materials and Information Engineering, 2022, VDE.

[4] Ali, A.O., Ahmed, A., Khan, A. and Patel, S. Closed-loop home energy management system with renewable energy sources in a smart grid: A comprehensive review. Journal of Energy Storage, 2022, 50, 104609.

[5] Wan, Q., Li, X., Zhao, Y. and Chen, H. Image anomaly detection and prediction scheme based on SSA optimized ResNet50-BiGRU model. arXiv preprint arXiv:2406.13987, 2024.

[6] Wang, C., Li, S., Zhao, M. and Zhang, R. A systematic review on power system resilience from the perspective of generation, network, and load. Renewable and Sustainable Energy Reviews, 2022, 167, 112567.

[7] Teekaraman, Y., Kumar, S. Patel, D., and Singh, R. SSNN-based energy management strategy in grid connected system for load scheduling and load sharing. Mathematical Problems in Engineering, 2022, 2022, 1-9.

[8] Ghasempour, A. and M. Martínez-Ramón. Short-term electric load prediction in smart grid using multi-output gaussian processes regression. in 2023 IEEE Kansas Power and Energy Conference (KPEC). 2023. IEEE.

[9] Hu, Y., et al. MultiLoad-GAN: A GAN-Based Synthetic Load Group Generation Method Considering Spatial-Temporal Correlations. IEEE Transactions on Smart Grid, 2023.

[10] Zhang, Y., et al. Sustainable Development of Smart Power Industry: Multi-level framework based on big data technology. Journal of Intelligence Technology and Innovation, 2024, 2, 38-59.

[11] Bashir, T., et al. Short term electricity load forecasting using hybrid prophet-LSTM model optimized by BPNN. Energy Reports, 2022, 8, 1678-1686.

[12] Li, L., Luo, D. and Yao, W. Analysis of transmission line icing prediction based on CNN and data mining technology. Soft Computing, 2022, 26(16), 7865-7870.

[13] Xue, X., et al. The Influence of Institutional Distance on the Reverse Technology Spillover Effect: Evidence from China. International Journal of Management and Organization, 2023, 1(1), 14-37.

[14] Sweeten, J., et al. Cyber-Physical GNN-Based Intrusion Detection in Smart Power Grids. in 2023 IEEE International Conference on Communications, Control, and Computing Technologies for Smart Grids (SmartGridComm). 2023. IEEE.

[15] Zhang, Q., et al. Fault diagnosis of power grid based on variational mode decomposition and convolutional neural network. Electric Power Systems Research, 2022, 208, 107871.

[16] Chung, W.H., Gu, Y.H. and Yoo, S.J. District heater load forecasting based on machine learning and parallel CNN-LSTM attention. Energy, 2022, 246, 123350.

[17] Akar, O., Terzi, U.K. and Ozgonenel, O. Location of transformers during the extension of an electricity distribution network. Electric Power Systems Research, 2022, 211, 108189.

[18] Takiddin, A., Ali, M., Zhang, H. and Liu, F. Deep autoencoder-based anomaly detection of electricity theft cyberattacks in smart grids. IEEE Systems Journal, 2022, 16(3), 4106-4117.

[19] Almalaq, A. and Edwards, G. A review of deep learning methods applied on load forecasting. in 2017 16th IEEE international conference on machine learning and applications (ICMLA). 2017. IEEE.

[20] Haq, M.R. and Ni, Z. A new hybrid model for short-term electricity load forecasting. IEEE access, 2019, 7, 125413-125423.

[21] Park, H., et al. Transformer network-based reinforcement learning method for power distribution network (PDN) optimization of high bandwidth memory (HBM). IEEE Transactions on Microwave Theory and Techniques, 2022, 70(11), 4772-4786.

[22] Li, G., Wang, Y., Zhang, X. and Chen, Z. Research on the Optimization of Air Route Network in Henan Province from the Perspective of Spatial Interaction Intensity-Taking the Four World-class Airport Clusters as Examples. Journal of Management Science and Operations, 2024, 2, 23-37.

[23] Wang, L., Zang, Y., Li, F. and Sun, Q. M2TNet: Multi-modal multi-task Transformer network for ultra-short-term wind power multi-step forecasting. Energy Reports, 2022, 8, 7628-7642.

[24] Huang, Y.-t., Chen, X., Zhang, W. and Liu, R. Application of a Hybrid Model Based on ICEEMDAN, Bayesian Hyperparameter Optimization GRU and the ARIMA in Nonferrous Metal Price Prediction. Cybernetics and Systems, 2023, 54(1), 27-59.

[25] ali Rostami, N. and Sadegh, M.O. The effect of load modeling on load flow results in distribution systems. American Journal of Electrical and Electronic Engineering, 2018, 6(1), 16-27.

[26] Liu, R., Wang, J, Li, Y. and Zhang, H. Artificial intelligence for fault diagnosis of rotating machinery: A review. Mechanical Systems and Signal Processing, 2018, 108, 33-47.

[27] Singh, K., Kumar, M, Joshi, R. and Sharma, P. An Optimal Parameterized Fractional-Order PID Controller for the Single Phase Grid Integrated with Solar and Wind System. Cybernetics and Systems, 2023, 54(7) 1086-1110.

[28] Mokarram, M.J., Ali, H., Reza, S. and Khan, Z. Net-load forecasting of renewable energy systems using multi-input LSTM fuzzy and discrete wavelet transform. Energy, 2023, 275, 127425.

[29] Cho, K., Bengio, Y. and LeCun, Y. Learning phrase representations using RNN encoder-decoder for statistical machine translation. arXiv preprint arXiv:1406.1078, 2014.

[30] Wang, J., Li, X., Zhang, Y., and Chen, Q. Towards Robust LiDAR-Camera Fusion in BEV Space via Mutual Deformable Attention and Temporal Aggregation. IEEE Transactions on Circuits and Systems for Video Technology, 2024.

[31] Bahdanau, D., Cho, K. and Bengio, Y. Neural machine translation by jointly learning to align and translate. arXiv preprint arXiv:1409.0473, 2014.

[32] Yu, Z., et al. MV-ReID: 3D Multi-view Transformation Network for Occluded Person Re-Identification. Knowledge-Based Systems, 2024, 283, 111200.

[33] Zhang, H., Liu, T. Wang, J. and Zhao, Q. PointGT: A Method for Point-Cloud Classification and Segmentation Based on Local Geometric Transformation. IEEE Transactions on Multimedia, 2024.

[34] Vaswani, A. et al. Attention is all you need. Advances in neural information processing systems, 2017. 30.

[35] Soydaner, D. Attention mechanism in neural networks: where it comes and where it goes. Neural Computing and Applications, 2022, 34(16), 13371-13385.

[36] Li, Y., Wang, Z., Zhang, T. and Chen, H. Deep learning based on Transformer architecture for power system short-term voltage stability assessment with class imbalance. Renewable and Sustainable Energy Reviews, 2024, 189, 113913.

[37] Dashtdar, M., Ahmadi, H., Taheri, R. and Ghasemi, S. Improving the power quality of island microgrid with voltage and frequency control based on a hybrid genetic algorithm and PSO. IEEE Access, 2022, 10, 105352-105365.

[38] Zhang, Q. and Li, F. A Dataset for Electricity Market Studies on Western and Northeastern Power Grids in the United States. Scientific Data, 2023, 10(1), 646.

[39] Khan, Z.A., Rehman, Z., Ahmed, S. and Hussain, F. Efficient short-term electricity load forecasting for effective energy management. Sustainable Energy Technologies and Assessments, 2022, 53, 102337.

[40] Zsiborács, H., et al. The accuracy of PV Power Plant Scheduling in Europe: An Overview of ENTSO-E Countries. IEEE Access, 2023.

[41] Alsharekh, M.F., Khalil, M.I., Jameel, M. and Ali, H. Improving the efficiency of multistep short-term electricity load forecasting via R-CNN with ML-LSTM. Sensors, 2022, 22(18), 6913.

[42] Ijaz, K., Hassan, M., Ahmad, F. and Sheikh, M.A. A novel temporal feature selection based LSTM model for electrical short-term load forecasting. IEEE Access, 2022, 10, 82596-82613.

[43] Chi, D. Research on electricity consumption forecasting model based on wavelet transform and multi-layer LSTM model. Energy Reports, 2022, 8, 220-228.

[44] Liserre, M., Blaabjerg, F. and Boldea, I. Unlocking the hidden capacity of the electrical grid through smart transformer and smart transmission. Proceedings of the IEEE, 2022.

[45] Huang, Z., Zhang, Y., Liu, F. and Ma, L. Multi-trajectory prediction of 5G network for smart grid based on Transformer. in 2022 4th International Conference on Communications, Information System and Computer Engineering (CISCE). 2022. IEEE.

[46] Zhang, D., Liu, J., Wang, P. and Chen, H. Real-time load forecasting model for the smart grid using bayesian optimized CNN-BiLSTM. Frontiers in Energy Research, 2023, 11, 1193662.

[47] Lin, C.-H., Chen, K., Huang, Y. and Liu, S. Nontechnical Loss Detection With Duffing–Holmes Self-Synchronization Dynamic Errors and 1D CNN-Based Multilayer Classifier in a Smart Grid. IEEE Access, 2022, 10, 83002-83016.

[48] Zhang, M., Zou, H., Farzamkia, S., Chen, Z. and Huang, A.Q., 2024, February. New Single-Stage Single-Phase Isolated Bidirectional AC-DC PFC Converter. In *2024 IEEE Applied Power Electronics Conference and Exposition (APEC)* (pp. 1962-1967). IEEE.

[49] Zou, H., Zhang, M., Farzamkia, S. and Huang, A.Q., 2024, February. Simplified Fixed Frequency Phase Shift Modulation for A Novel Single-Stage Single Phase Series-Resonant AC-DC Converter. In *2024 IEEE Applied Power Electronics Conference and Exposition (APEC)* (pp. 1261-1268). IEEE.

[50] Zhang, M., Zou, H., Farzamkia, S., Chen, Z. and Huang, A.Q., 2024, February. New Single-Stage Single-Phase Isolated Bidirectional AC-DC PFC Converter. In *2024 IEEE Applied Power Electronics Conference and Exposition (APEC)* (pp. 1962-1967). IEEE.

[51] Xie, X., Peng, H., Hasan, A., Huang, S., Zhao, J., Fang, H., Zhang, W., Geng, T., Khan, O., & Ding, C. (2023). Accel-gcn: High-performance gpu accelerator design for graph convolution networks. In 2023 IEEE/ACM International Conference on Computer Aided Design (ICCAD) (pp. 01-09). IEEE.

[52] Liu, Y., Zhao, R., & Li, Y. (2022). A Preliminary Comparison of Drivers' Overtaking behavior between Partially Automated Vehicles and Conventional Vehicles. In Proceedings of the Human Factors and Ergonomics Society Annual Meeting, 66(1), 913-917.

[53] Xu, Z., Deng, D., Dong, Y., & Shimada, K. (2022). DPMPC-Planner: A real-time UAV trajectory planning framework for complex static environments with dynamic obstacles. In 2022 International Conference on Robotics and Automation (ICRA), 250-256.

[54] Zhao, R., Liu, Y., Li, T., & Li, Y. (2022). A Preliminary Evaluation of Driver's Workload in Partially Automated Vehicles. In International Conference on Human-Computer Interaction, 448-458.

[55] Richardson, A., Wang, X., Dubey, A., & Sprinkle, J. (2024). Reinforcement Learning with Communication Latency with Application to Stop-and-Go Wave Dissipation. In 2024 IEEE Intelligent Vehicles Symposium (IV), 1187-1193.

[56] Peng, H., Huang, S., Zhou, T., Luo, Y., Wang, C., Wang, Z., Zhao, J., Xie, X., Li, A., Geng, T., & others. (2023). AutoReP: Automatic ReLU Replacement for Fast Private Network Inference. In 2023 IEEE/CVF International Conference on Computer Vision (ICCV) (pp. 5155-5165). IEEE.

[57] Song, Y., Fellegara, R., Iuricich, F., & De Floriani, L. (2024). Parallel Topology-aware Mesh Simplification on Terrain Trees. ACM Transactions on Spatial Algorithms and Systems, 10(2), 1-39.

[58] De, A., Mohammad, H., Wang, Y., Kubendran, R., Das, A. K., & Anantram, M. P. (2023). Performance analysis of DNA crossbar arrays for high-density memory storage applications. Scientific Reports, 13(1), 6650.

[59] Jin, C., Peng, H., Zhao, S., Wang, Z., Xu, W., Han, L., Zhao, J., Zhong, K., Rajasekaran, S., & Metaxas, D. N. (2024). APEER: Automatic Prompt Engineering Enhances Large Language Model Reranking. arXiv preprint arXiv:2406.14449.

[60] Patel, S., Liu, Y., Zhao, R., Liu, X., & Li, Y. (2022). Inspection of in-vehicle touchscreen infotainment display for different screen locations, menu types, and positions. In International Conference on Human-Computer Interaction, 258-279.

[61] Liu, Y., Zhao, R., Li, T., & Li, Y. (2022). The Impact of Directional Road Signs Combinations and Language Unfamiliarity on Driving Behavior. In International Conference on Human-Computer Interaction, 195-204.

[62] De, A., Mohammad, H., Wang, Y., Kubendran, R., Das, A. K., & Anantram, M. P. (2022). Modeling and Simulation of DNA Origami based Electronic Read-only Memory. In 2022 IEEE 22nd International Conference on Nanotechnology (NANO), 385-388.

[63] Dong, Y. (2024). The Design of Autonomous UAV Prototypes for Inspecting Tunnel Construction Environment. arXiv preprint arXiv:2408.07286.

[64] Wang, Y., Alangari, M., Hihath, J., Das, A. K., & Anantram, M. P. (2021). A machine learning approach for accurate and real-time DNA sequence identification. BMC Genomics, 22, 1-10.

[65] Li, T., Zhao, R., Liu, Y., Liu, X., & Li, Y. (2022). Effect of Age on Driving Behavior and a Neurophysiological Interpretation. In International Conference on Human-Computer Interaction, 184-194.

[66] Luo, Y., Xu, N., Peng, H., Wang, C., Duan, S., Mahmood, K., Wen, W., Ding, C., & Xu, X. (2023). AQ2PNN: Enabling Two-party Privacy-Preserving Deep Neural Network Inference with Adaptive Quantization. In 2023 56th IEEE/ACM International Symposium on Microarchitecture (MICRO) (pp. 628-640). IEEE.

[67] Li, T., Zhao, R., Liu, Y., Li, Y., & Li, G. (2021). Evaluate the effect of age and driving experience on driving performance with automated vehicles. In International Conference on Applied Human Factors and Ergonomics, 155-161.

[68] Wang, S., Jiang, R., Wang, Z., & Zhou, Y. (2024). Deep Learning-based Anomaly Detection and Log Analysis for Computer Networks. Journal of Information and Computing, 2(2), 34-63.

[69] Peng, X., Xu, Q., Feng, Z., Zhao, H., Tan, L., Zhou, Y., Zhang, Z., Gong, C., & Zheng, Y. (2024). Automatic News Generation and Fact-Checking System Based on Language Processing. arXiv preprint arXiv:2405.10492.

[70] Zhu, Z., Zhao, R., Ni, J., & Zhang, J. (2019). Image and spectrum based deep feature analysis for particle matter estimation with weather information. In 2019 IEEE International Conference on Image Processing (ICIP), 3427-3431.

[71] Wan, Q., Zhang, Z., Jiang, L., Wang, Z., & Zhou, Y. (2024). Image anomaly detection and prediction scheme based on SSA optimized ResNet50-BiGRU model. arXiv preprint arXiv:2406.13987.

[72] Deng, Q., Fan, Z., Li, Z., Pan, X., Kang, Q., & Zhou, M. (2024). Solving the Food-Energy-Water Nexus Problem via Intelligent Optimization Algorithms. arXiv preprint arXiv:2404.06769.

[73] Peng, H., Ran, R., Luo, Y., Zhao, J., Huang, S., Thorat, K., Geng, T., Wang, C., Xu, X., Wen, W., & others. LinGCN: Structural Linearized Graph Convolutional Network for Homomorphically Encrypted Inference. In Thirty-seventh Conference on Neural Information Processing Systems.

[74] Chen, P., Zhang, Z., Dong, Y., Zhou, L., & Wang, H. (2024). Enhancing Visual Question Answering through Ranking-Based Hybrid Training and Multimodal Fusion. arXiv preprint arXiv:2408.07303.

[75] Jiang, X., Yu, J., Qin, Z., Zhuang, Y., Zhang, X., Hu, Y., & Wu, Q. (2020). Dualvd: An adaptive dual encoding model for deep visual understanding in visual dialogue. In Proceedings of the AAAI Conference on Artificial Intelligence, 34(07), 11125-11132.

[76] Wang, X., Onwumelu, S., & Sprinkle, J. (2024). Using Automated Vehicle Data as a Fitness Tracker for Sustainability. In 2024 Forum for Innovative Sustainable Transportation Systems (FISTS), 1-6.

[77] Jiang, L., Yu, C., Wu, Z., Wang, Y., & others. (2024). Advanced AI Framework for Enhanced Detection and Assessment of Abdominal Trauma: Integrating 3D Segmentation with 2D CNN and RNN Models. arXiv preprint arXiv:2407.16165.

[78] Wang, C., Sui, M., Sun, D., Zhang, Z., & Zhou, Y. (2024). Theoretical Analysis of Meta Reinforcement Learning: Generalization Bounds and Convergence Guarantees. arXiv preprint arXiv:2405.13290.

[79] Peng, H., Xie, X., Shivdikar, K., Hasan, M. A., Zhao, J., Huang, S., Khan, O., Kaeli, D., & Ding, C. (2024). MaxK-GNN: Extremely Fast GPU Kernel Design for Accelerating Graph Neural Networks Training. In Proceedings of the 29th ACM International Conference on Architectural Support for Programming Languages and Operating Systems, Volume 2 (pp. 683-698). Association for Computing Machinery, New York, NY, USA.

[80] Zhao, R., Liu, Y., Li, Y., & Tokgoz, B. (2021). An Investigation of Resilience in Manual Driving and Automatic Driving in Freight Transportation System. In IIE Annual Conference. Proceedings, 974-979.

[81] Fellegara, R., Iuricich, F., Song, Y., & Floriani, L. D. (2023). Terrain trees: A framework for representing, analyzing and visualizing triangulated terrains. GeoInformatica, 27(3), 525-564.

[82] Wang, Y., Khandelwal, V., Das, A. K., & Anantram, M. P. (2022). Classification of DNA Sequences: Performance Evaluation of Multiple Machine Learning Methods. In 2022 IEEE 22nd International Conference on Nanotechnology (NANO), 333-336.

[83] Zhou, T., Zhao, J., Luo, Y., Xie, X., Wen, W., Ding, C., & Xu, X. (2024). AdaPI: Facilitating DNN Model Adaptivity for Efficient Private Inference in Edge Computing. arXiv preprint arXiv:2407.05633.

[84] Jin, C., Huang, T., Zhang, Y., Pechenizkiy, M., Liu, S., Liu, S., & Chen, T. (2023). Visual prompting upgrades neural network sparsification: A data-model perspective. arXiv preprint arXiv:2312.01397.

[85] Wang, Y., Demir, B., Mohammad, H., Oren, E. E., & Anantram, M. P. (2023). Computational study of the role of counterions and solvent dielectric in determining the conductance of B-DNA. Physical Review E, 107(4), 044404.

[86] Lee, J. W., Wang, H., Jang, K., Hayat, A., Bunting, M., Alanqary, A., Barbour, W., Fu, Z., Gong, X., Gunter, G., & others. (2024). Traffic smoothing via connected & automated vehicles: A modular, hierarchical control design deployed in a 100-cav flow smoothing experiment. IEEE Control Systems Magazine.

[87] Zhou, Y., Wang, Z., Zheng, S., Zhou, L., Dai, L., Luo, H., Zhang, Z., & Sui, M. (2024). Optimization of automated garbage recognition model based on ResNet-50 and weakly supervised CNN for sustainable urban development. Alexandria Engineering Journal, 108, 415-427.

[88] Liu, Y., Zhao, R., Li, T., & Li, Y. (2021). An investigation of the impact of autonomous driving on driving behavior in traffic jam. In IIE Annual Conference. Proceedings, 986-991.

[89] An, Z., Wang, X., T. Johnson, T., Sprinkle, J., & Ma, M. (2023). Runtime monitoring of accidents in driving recordings with multi-type logic in empirical models. In International Conference on Runtime Verification, 376-388.

[90] Zhuang, Y., Chen, Y., & Zheng, J. (2020). Music genre classification with transformer classifier. In Proceedings of the 2020 4th International Conference on Digital Signal Processing, 155-159.

[91] Jin, C., Che, T., Peng, H., Li, Y., & Pavone, M. (2024). Learning from teaching regularization: Generalizable correlations should be easy to imitate. arXiv preprint arXiv:2402.02769.

[92] Song, Y., Fellegara, R., Iuricich, F., & De Floriani, L. (2021). Efficient topology-aware simplification of large triangulated terrains. In Proceedings of the 29th International Conference on Advances in Geographic Information Systems, 576-587.